\documentclass[conference]{IEEEtran}
\IEEEoverridecommandlockouts
\usepackage{cite}
\usepackage{amsmath,amssymb,amsfonts}
\usepackage{algorithmic}
\usepackage{graphicx}
\usepackage{textcomp}
\usepackage{xcolor}
\usepackage{gensymb}
\usepackage{url}
\def\BibTeX{{\rm B\kern-.05em{\sc i\kern-.025em b}\kern-.08em
    T\kern-.1667em\lower.7ex\hbox{E}\kern-.125emX}}
\begin{document}

\title{Critic Architecture Matters: Dual vs.\ Unified Critics for Humanoid Loco-Manipulation}

\author{\IEEEauthorblockN{Mehmet Turan Yardımcı}
\IEEEauthorblockA{\textit{Department of Computer Engineering}\\
\textit{Çukurova University}\\
Adana, Türkiye\\
mehmetturan2003@gmail.com}
}

\maketitle

\begin{abstract}
Multi-objective reinforcement learning for humanoid robots must coordinate locomotion and manipulation within one policy. A natural design choice is between a single (unified) critic that estimates the combined value of all objectives and separate (dual) critics with disjoint reward signals. We compare the two on the Unitree G1 humanoid in NVIDIA Isaac Lab. In the standing mode of a standardized evaluation, the dual-critic run reaches targets 3.5$\times$ faster (6.5 vs.\ 22.6 simulation steps), achieves 2$\times$ the throughput (14.3 vs.\ 7.0 validated reaches per 1{,}000 steps) and a higher validated reach rate (65.2\% vs.\ 53.8\%) than the unified-critic run. That evaluation pins the fingers open for every policy, whereas the unified run had trained driving its own. When the unified run drives its own fingers, with nothing else changed, the standing-mode gap falls from 3.5$\times$ to 1.3$\times$ in speed and from 2$\times$ to 1.1$\times$ in throughput. This is a single re-evaluation of a single checkpoint, and we do not generalize from it. Adding five anti-gaming reward mechanisms to the dual critic did not raise validated reach rate (60.9\% vs.\ 65.2\%). The two runs differ not only in the critic but also in the PPO update rule (one summed advantage under one likelihood ratio, versus a per-stream advantage and a ratio per actor), and further in curriculum, arm action dimensionality, finger control and reward weights; each is a single run. The measurement therefore cannot separate the critic from the update rule. We argue that critic architecture deserves explicit treatment as a design variable in multi-objective humanoid RL, and specify the single-variable ablation needed to establish its causal contribution. Code, checkpoints and a project page: \url{https://mturan33.github.io/critic-architecture-matters/}
\end{abstract}


\section{Introduction}

Humanoid loco-manipulation---the simultaneous coordination of walking and reaching---is a fundamental capability for robots operating in human environments. While imitation learning approaches such as ALOHA~\cite{zhao2023aloha} and Mobile ALOHA~\cite{fu2024mobilealoha} have demonstrated impressive manipulation, reinforcement learning (RL) offers a complementary paradigm. Consider how infants learn to walk: they begin by crawling---with no need to observe other crawlers---then progress to assisted walking, and eventually walk independently, in a process analogous to curriculum learning. This exploration-driven process stands in direct contrast to the imitation learning paradigm. Yet these paradigms are increasingly combined: recent work refines imitation-learned policies with RL to improve robustness and generalization beyond the demonstration distribution~\cite{haldar2023teach}. In such hybrid pipelines, the critic architecture plausibly governs whether RL gradients \textit{refine} or \textit{overwrite} the pre-trained behavior---a question this paper motivates but does not test. However, RL introduces its own challenges, particularly when multiple objectives must be coordinated within a single policy.

A fundamental design choice in multi-objective RL is the critic architecture. A \textbf{unified critic} receives the concatenated observations from all objectives and estimates their combined value, while \textbf{dual critics} maintain separate value functions with disjoint reward signals. Separate critics per reward group are already used in legged and humanoid RL and have been ablated against a single critic~\cite{vijayan2025,robotkeyframing,beamdojo,cwi}, including a double-critic ablation on the Unitree G1 in BeamDojo~\cite{beamdojo} and, in humanoid loco-manipulation, CWI~\cite{cwi}, where replacing three critics with one raised end-effector tracking error in simulation. This paper adds a comparison of the two designs on the Unitree G1, with an account of what a single pair of runs can and cannot establish.

Both runs learn functional reaching, and they differ in \textit{efficiency} by an amount that depends on the evaluation protocol: in the standing mode of our original, pinned-finger evaluation the dual-critic policy reaches 3.5$\times$ faster and achieves 2$\times$ higher throughput, figures that fall to 1.3$\times$ and 1.1$\times$ when the unified policy drives its own fingers. Crucially, the training metrics one would normally consult cannot show this difference: final reward reads 36.2 against 37.1, but the runs do not share a reward table (Sec.~\ref{sec:confounds}), so the two numbers are not on a common scale and their closeness says nothing about comparable performance. Standardized evaluation with validated metrics is what makes the difference visible at all; how large it looks also depends on that evaluation, as Sec.~\ref{sec:results} shows.

The contributions of this paper are: (1)~A Dual Actor-Critic framework with frozen-branch transfer for humanoid loco-manipulation on the Unitree G1 in NVIDIA Isaac Lab. (2)~A three-way benchmark in which the dual-critic run reaches 3.5$\times$ faster and attains 2$\times$ higher throughput than the unified-critic run in the standing mode (2.3$\times$ and 1.8$\times$ walking); with the unified run driving its own fingers the figures are 1.3$\times$ and 1.1$\times$ standing (1.2$\times$ and 1.03$\times$ walking). Five additional anti-gaming reward mechanisms did not raise validated reach rate in this run. (3)~A standardized benchmark methodology: the two runs' training curves cannot be compared, because their reward tables and reach definitions differ, and the difference is measurable only in the standardized evaluation. We are explicit about what this design does \textit{not} establish: the two runs differ in more than the critic (Sec.~\ref{sec:confounds}), so the efficiency gap is reported as an observation rather than an isolated causal effect.

\section{Related Work}

\textbf{Humanoid Locomotion and Loco-Manipulation.}
Radosavovic et al.~\cite{radosavovic2024humanoid} demonstrated zero-shot sim-to-real transfer of a locomotion policy on a full-size humanoid. He et al.~\cite{he2024hover} proposed HOVER for multi-mode whole-body control, and Sun et al.~\cite{sun2025ulc} introduced ULC, a unified controller for the G1 using sequential skill acquisition. Fu et al.~\cite{fu2022deep} and Cheng et al.~\cite{cheng2024expressive} demonstrated whole-body control for legged manipulation. These works adopt specific architectures without comparing alternatives. Our work treats the critic architecture as a variable to be measured rather than assumed, and reports the resulting efficiency difference together with the factors that co-vary with it.

\textbf{Curriculum Learning in RL.}
Curriculum learning~\cite{bengio2009curriculum} proposes training on progressively harder tasks. OpenAI's Rubik's Cube work~\cite{openai2019rubiks} used automatic domain randomization as an implicit curriculum, and Narvekar et al.~\cite{narvekar2020curriculum} surveyed curriculum methods. ULC~\cite{sun2025ulc} employs sequential skill acquisition similar to ours. Our experience adds a practical caution: curriculum level is only interpretable relative to the schedule that produced it, and comparing level counters across runs with different schedules is misleading.

\textbf{Multi-Objective RL and Critic Design.}
Reward hacking~\cite{amodei2016concrete,skalse2022defining,pan2022effects} is well-documented, but typically attributed to reward design. Multi-critic approaches are established in multi-agent settings, and single-agent robotics has adopted them as well, training a separate critic per reward group in legged loco-manipulation~\cite{vijayan2025}, quadruped locomotion~\cite{robotkeyframing}, humanoid locomotion~\cite{beamdojo} and humanoid loco-manipulation~\cite{cwi}; each of these works also compares its multi-critic design against a single critic, as one component of the proposed method. Our work contributes an empirical data point on separating critics by objective in humanoid control, together with an explicit account of what a two-run comparison can and cannot establish.

\textbf{RL Fine-Tuning of Imitation-Learned Policies.}
A growing body of work uses RL to refine policies initially trained via imitation learning. Haldar et al.~\cite{haldar2023teach} keep an imitation-learned base policy fixed and train an RL residual on it, adapting it to object configurations not seen in the demonstrations, while Luo et al.~\cite{luo2024serl} released a software suite for sample-efficient real-world RL that draws half of each update batch from demonstrations rather than starting from an imitation-learned policy. Refining an imitation-learned policy with RL faces a tension: the RL objective must improve specific behaviors without catastrophically forgetting the imitation-learned skills. This tension is structurally analogous to our multi-objective setting, where locomotion and manipulation compete for gradient signal through the critic. Whether separate critics reduce that interference, and would thereby protect IL-learned behaviors during RL fine-tuning, is an open question that our experiments do not test.

\section{Method}

\subsection{System Overview}

We train a Unitree G1 humanoid robot in NVIDIA Isaac Lab~\cite{mittal2023orbit}. The robot has 23 active body DoF, of which the policy controls 17 in every run: 12 leg joints and 5 arm joints (right arm: shoulder pitch/roll/yaw, elbow pitch/roll). Wrist joints are held fixed throughout. The seven right-hand Dex3 finger joints lie outside that count of 23 and are not a fixed part of the control surface: they are pinned open in the dual-critic runs and in the original, pinned-finger evaluation, but the unified run drove them from Level 10 onward, so its policy commanded 24 joints for most of its training (Sec.~\ref{sec:confounds}). Parallel environment counts differ by run and are reported per run in Sec.~\ref{sec:setup}.

Training proceeds sequentially: Stages 1--3 train locomotion only; Stage~6 trains both branches jointly for loco-manipulation. Both branches use PPO~\cite{schulman2017ppo} with learning rate $3 \times 10^{-4}$ (cosine annealing), $\gamma=0.99$, $\lambda=0.95$, and clip ratio 0.2.

\textbf{Unified Critic Architecture.} Our initial design uses separate actors $\pi_{\text{loco}}$ (57-dim obs $\to$ 12 leg actions) and $\pi_{\text{arm}}$ (52-dim obs $\to$ 12 arm+finger actions), but a \textbf{single critic} $V_{\text{unified}}$ that receives the concatenated 109-dimensional observation and estimates the combined value of locomotion and manipulation rewards.

\begin{figure}[t]
\centering
\includegraphics[width=\columnwidth]{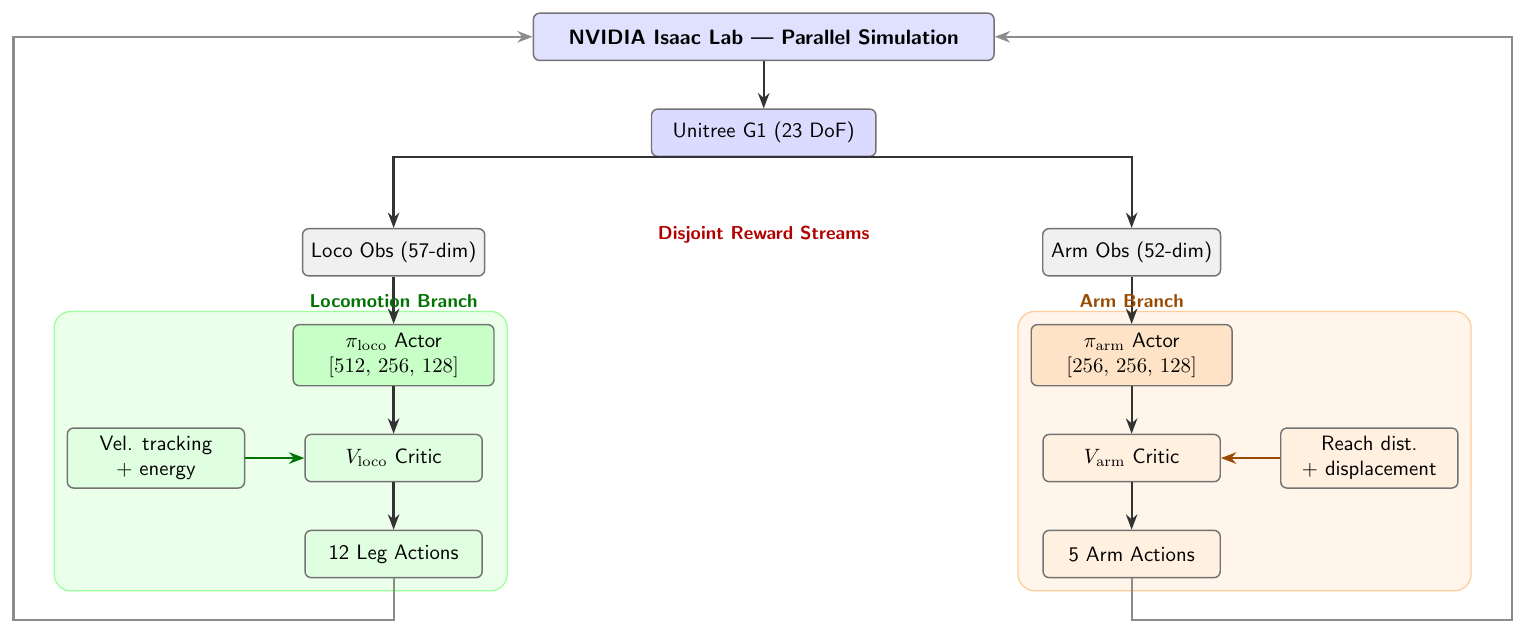}
\caption{Dual Actor-Critic architecture. The locomotion branch (green) and arm branch (orange) maintain separate critics with disjoint reward streams. Observation widths are those of the main comparison (S6s); the S7 variant adds three anti-gaming features to the arm branch (55-dim) and freezes the locomotion branch. The unified variant (not shown) concatenates both observation streams into a single 109-dim critic input.}
\label{fig:architecture}
\end{figure}

\textbf{Dual Actor-Critic Architecture.} The revised design (Fig.~\ref{fig:architecture}) separates the system into fully independent branches. The \textit{Locomotion branch} (actor $\pi_{\text{loco}}$, critic $V_{\text{loco}}$) observes proprioceptive state (57-dim: base velocities, projected gravity, leg joint states, velocity commands, gait phase) and outputs 12 leg joint targets. The \textit{Arm branch} (actor $\pi_{\text{arm}}$, critic $V_{\text{arm}}$) observes arm-relevant state (52-dim: base motion, leg context,
arm joint states, end-effector/target positions; 55-dim in S7 with 3 anti-gaming features) and outputs 5 arm joint residual actions. The two critics receive \textbf{disjoint reward signals}: locomotion critic on velocity tracking and balance, arm critic on reaching distance and displacement only.

\subsection{Sequential Curriculum Design}

\textbf{The two runs follow different curricula, and this matters for how the results should be read.} The dual-critic run (S6s) uses the 13-level schedule of Table~\ref{tab:curriculum}: stationary reaching (Levels 0--4), walking with reaching (5--6), fixed end-effector orientation (7--8), and variable orientation within an expanding cone (9--12, 20\degree--80\degree). The unified-critic run (S6u) predates this schedule and uses a 40-level one organized on different axes: reaching only (0--9), $+$orientation \emph{and} gripper together (10--19), $+$commanded torso height (20--29), and $+$payload (30--39). It is not a finer-grained version of the same ladder---it resets the commanded base velocity to zero at each phase boundary and re-ramps, it never trains variable end-effector orientation, and it trains gripper control and payload handling, which the dual-critic run never sees. The unified run entered its gripper phase: it reached Level 10 at iteration 1{,}162 and spent 18{,}838 of its 20{,}000 iterations there. Advancement in the unified run requires a sustained ratio of reaches to steps, in the dual run a sustained validated reach rate, and the thresholds differ.

Consequently the two curriculum-level counters are not on a common scale, and the task each policy was training on when its checkpoint was taken differs substantially. At its final level the unified policy was commanded to \textit{stand still} and reach targets sampled 0.18--0.35\,m from the shoulder within 0.05\,m, palm roughly downward (0.8\,rad tolerance) with gripper control active; the dual policy was \textit{walking at up to 0.6\,m/s} and reaching targets 0.18--0.40\,m from the shoulder within 0.04\,m at a commanded palm direction sampled from an 80\degree\ cone (1.0\,rad tolerance). We return to this in Sec.~\ref{sec:confounds}.

\begin{table}[h]
\centering
\caption{Curriculum levels for the dual-critic run (S6s). The unified-critic run (S6u) uses a different 40-level schedule; see text.}
\label{tab:curriculum}
\begin{tabular}{|c|c|c|c|c|}
\hline
\textbf{Level} & \textbf{Phase} & \textbf{$v_x$ (m/s)} & \textbf{Pos. Thresh.} & \textbf{Orient.} \\
\hline
0--4 & Stand+Reach & 0.0 & 0.12$\to$0.06m & None \\
5--6 & Walk+Reach & 0--0.3 & 0.06$\to$0.05m & None \\
7--8 & Fixed Orient & 0--0.4 & 0.05m & Palm-down \\
9--12 & Var. Orient & 0--0.6 & 0.05$\to$0.04m & 20--80\degree \\
\hline
\end{tabular}
\end{table}

\subsection{Anti-Gaming Reward Mechanisms}

During development, we observed that proximity-based rewards combined with automatic target resampling could allow policies to accumulate spurious reach counts. To investigate whether reward engineering could further improve performance beyond the architectural change, we implemented five anti-gaming mechanisms in a third variant (Stage~7): (1)~absolute workspace sampling with minimum distance constraints, (2)~three-condition validated reach requiring position proximity, displacement, and time limit:
\begin{equation}
\text{valid} = (\|p_{ee} - p_t\| < \epsilon_{pos}) \wedge (\|p_{ee} - p_{ee}^{0}\| > d_{disp}) \wedge (t < t_{max})
\end{equation}
(3)~movement-centric rewards with velocity-toward-target and stillness penalties, (4)~validated reach rate for curriculum advancement, and (5)~gaming detection heuristics. This variant uses the dual-critic architecture with a frozen locomotion branch and freshly initialized arm policy. It also runs its own 8-level curriculum (three phases: standing, walking, walking with fixed palm-down orientation), distinct again from the two schedules above.

\section{Experiments}

\subsection{Setup}
\label{sec:setup}

All runs use NVIDIA Isaac Lab on a single RTX 5070 Ti GPU (12GB VRAM), achieving approximately 17{,}000 steps/second. The two runs of the main comparison (S6u, S6s) each used 2048 parallel environments for 20{,}000 iterations; the anti-gaming variant (S7) used 4096 for 15{,}000 iterations. We compare three policies in a standardized benchmark with identical evaluation conditions (absolute target sampling, minimum target distance 0.12m, position threshold 0.06m, displacement threshold 0.10m, timeout 150 steps, 1 environment, deterministic actions, seed 42):

\begin{itemize}
\item \textbf{S6u} (Unified Critic): Unified critic (109-dim), 52-dim arm obs, 12-dim arm action, final level 10 of 40
\item \textbf{S6s} (Dual Critic): Dual Actor-Critic, 52-dim arm obs, 5-dim arm action, final level 12 of 13
\item \textbf{S7} (Dual + Anti-Gaming): Dual Actor-Critic with anti-gaming reward mechanisms, frozen locomotion, fresh arm policy, final level 7 of 8
\end{itemize}

Level indices are zero-based and each is relative to that run's own schedule, so the final index of a completed curriculum is one less than its length: S6s and S7 reached the terminal level of theirs, while S6u stopped a quarter of the way into a longer, differently organized one.\footnote{The launch configuration of the S6u run was not recorded. Its environment count (2048) and iteration budget (20{,}000) are reconstructed from run artifacts---the saved checkpoints and the per-iteration reach counter, whose ceiling of $\text{num\_envs}\times24$ excludes 1024 and is never exceeded at 2048. We flag this explicitly because the auditability we advocate below should apply to our own data; all subsequent runs emit a machine-readable configuration record.}

Each policy is evaluated over 3{,}000 steps in both standing and walking modes. The dual and anti-gaming runs carry bit-identical locomotion weights, which the harness verifies at load time. The unified run's own locomotion weights differ from them, and the harness replaces that branch with the shared one, so the unified arm policy is evaluated with a locomotion branch it did not train with.

\subsection{Results}
\label{sec:results}

Table~\ref{tab:results} presents the three-way comparison. All three policies successfully reach targets; they differ in efficiency, by an amount that depends on the protocol. One caveat belongs with the table itself: S6u drove its own finger joints for 18{,}838 of its 20{,}000 training iterations, whereas the original, pinned-finger evaluation pins the fingers open for every policy, so the S6u column is measured in a configuration that run did not train in. We measured what this costs. Re-evaluating the same S6u checkpoint with its own finger outputs driving the fingers, changing nothing else, raises its validated reach rate from 53.8\% to 59.4\% standing and from 51.1\% to 62.5\% walking, cuts its standing-mode mean time-to-reach from 22.6 to 8.3 steps, and raises its standing-mode throughput from 7.0 to 12.7 validated reaches per 1{,}000 steps. With S6u driving its own fingers and nothing else changed, the standing-mode gap falls from 3.5$\times$ to 1.3$\times$ in speed and from 2$\times$ to 1.1$\times$ in throughput. Table~\ref{tab:results} keeps the original pinned-finger measurement and reports the re-evaluation beside it. This is a single re-evaluation of a single checkpoint, and we do not generalize from it. It also leaves a second mismatch in place: S6u still runs on the locomotion branch shared by the dual-critic runs rather than on its own, and we have not measured what that costs.

\begin{table}[h]
\centering
\caption{Standardized benchmark (3{,}000 steps per mode). S6u: unified critic. S6s: dual critic. S7: dual critic + anti-gaming.}
\label{tab:results}
\begin{tabular}{|l|c|c|c|}
\hline
\textbf{Metric} & \textbf{S6u} & \textbf{S6s} & \textbf{S7} \\
\hline
Critic architecture & Unified & Dual & Dual+AG \\
Final level$^{*}$ & 10/40 & 12/13 & 7/8 \\
Parallel environments & 2048 & 2048 & 4096 \\
Training iterations & 20{,}000 & 20{,}000 & 15{,}000 \\
\hline
\textit{Standing evaluation:} & & & \\
Validated reach rate & 53.8\% & 65.2\% & 60.9\% \\
Position-only rate & 59.0\% & 72.7\% & 71.9\% \\
Avg. time-to-reach & 22.6 steps & 6.5 steps & 5.8 steps \\
Validated / 1K steps & 7.0 & 14.3 & 13.0 \\
Timeout rate & 41.0\% & 27.3\% & 28.1\% \\
Avg. displacement & 0.219m & 0.206m & 0.203m \\
\hline
\textit{Walking evaluation:} & & & \\
Validated reach rate & 51.1\% & 63.1\% & 56.2\% \\
Avg. time-to-reach & 16.4 steps & 7.0 steps & 5.9 steps \\
Validated / 1K steps & 7.7 & 13.7 & 12.0 \\
Timeout rate & 37.8\% & 26.2\% & 28.1\% \\
\hline
Arm action magnitude$^{\dagger}$ & 2.43 & 3.35 & 2.81 \\
\hline
\multicolumn{4}{l}{\textit{S6u re-evaluated driving its own fingers (standing / walking):$^{\ddagger}$}} \\
Validated reach rate & 59.4\% / 62.5\% & --- & --- \\
Avg. time-to-reach & 8.3 / 8.2 steps & --- & --- \\
Validated / 1K steps & 12.7 / 13.3 & --- & --- \\
Timeout rate & 26.6\% / 26.6\% & --- & --- \\
Arm action magnitude$^{\dagger}$ & 2.14 / 2.12 & --- & --- \\
\hline
\end{tabular}

\vspace{2pt}
{\footnotesize $^{*}$Zero-based final level index over the length of that run's own schedule; S6s and S7 reached their terminal level. The three schedules are not on a common scale (Sec.~\ref{sec:confounds}).}

{\footnotesize $^{\ddagger}$Same checkpoint, same benchmark, same seed and same number of steps; the only change is that S6u's seven finger outputs drive the finger joints, as they did during training, instead of being discarded. S6s and S7 have no finger outputs and are unaffected, so those cells are empty rather than zero.}

{\footnotesize $^{\dagger}$Mean absolute arm action over the five arm dimensions, averaged across all 3{,}000 steps of the standing evaluation (of each mode in the $^{\ddagger}$ block). In v1 and v2 this row instead held 1.22, 2.54 and 3.04, which a debug print in the benchmark had produced from the \emph{first step alone}; that output was never written to a file. Re-running the same checkpoints reproduces those three values exactly, and also shows that one step misorders the two dual-critic runs (it puts S7 above S6s, the session means put S6s above S7). The row now reports the session means.}
\end{table}

Under the original protocol the largest ratio is in \textbf{reaching speed}: the unified-critic policy requires 22.6 simulation steps (0.45s) to reach a target on average, while the dual-critic policy requires only 6.5 steps (0.13s)---a 3.5$\times$ improvement in the standing mode. \textbf{Throughput} differs less than speed in the standing mode: dual critics achieve 14.3 validated reaches per 1{,}000 steps compared to 7.0 for the unified critic, a 2$\times$ difference. With S6u driving its own fingers the two ratios become 1.3$\times$ and 1.1$\times$. In the walking mode the original ratios are 2.3$\times$ in speed (16.4 vs.\ 7.0 steps) and 1.8$\times$ in throughput (13.7 vs.\ 7.7), and with S6u driving its own fingers they become 1.2$\times$ (8.2 vs.\ 7.0 steps) and 1.03$\times$ (13.7 vs.\ 13.3).

One observation accompanies the table rather than explains it. Averaged over the full standing evaluation, the unified-critic policy commands arm actions of mean magnitude 2.43 against 3.35 and 2.81 for the dual-critic policies---about three quarters of S6s, not the half that the first-step numbers of v1 and v2 suggested. Driving its own fingers lowers it further, to 2.14, while improving every reach metric, so smaller commanded arm actions do not track worse reaching even within this one policy. We therefore record the magnitudes as an observation only.

\begin{figure}[t]
\centering
\includegraphics[width=\columnwidth]{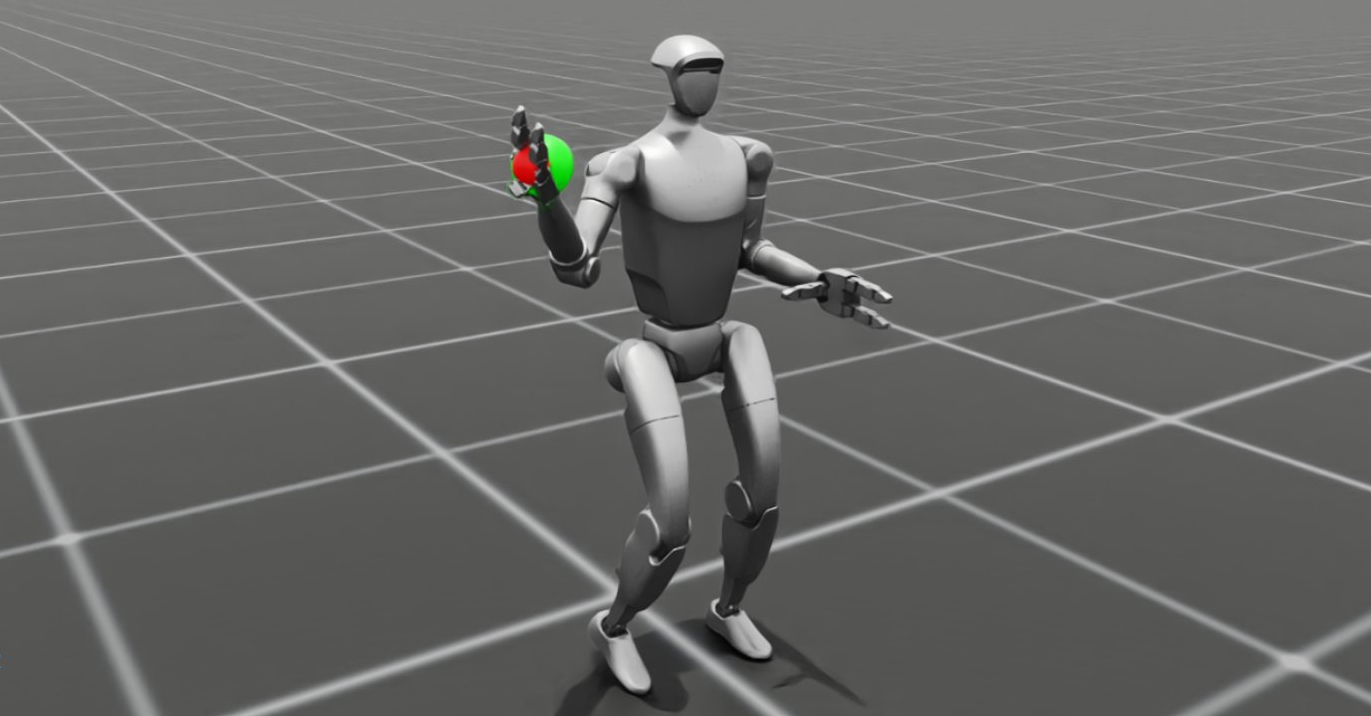}
\caption{Dual-critic policy (S6s) during play evaluation. The Unitree G1 reaches toward a target (green sphere) with the red marker indicating the end-effector tracking point.}
\label{fig:reaching}
\end{figure}

Fig.~\ref{fig:reaching} shows the trained dual-critic policy during play evaluation, demonstrating active reaching toward targets during locomotion.

\subsection{Discussion}

Three findings emerge from the benchmark, followed by one open question:

\textbf{(1) The two runs differ in standing-mode efficiency, by 3.5$\times$ in speed with S6u's fingers pinned and 1.3$\times$ with them driven.} Under the original protocol the dual-critic run is 3.5$\times$ faster to target, delivers 2$\times$ the throughput, and lands 11 percentage points higher on validated rate (65.2\% vs.\ 53.8\%). With S6u driving its own fingers and nothing else changed, the standing-mode figures become 1.3$\times$, 1.1$\times$ and 5.8 percentage points. One open hypothesis is that a unified critic, estimating a combined value across competing objectives, lets locomotion reward suppress arm action magnitudes; our own action-magnitude data do not support it, since S6u commands still smaller arm actions yet reaches better once its fingers are driven (Table~\ref{tab:results}). As Sec.~\ref{sec:confounds} details, the two runs also differ in their update rule---one summed advantage under a joint likelihood ratio against a per-stream advantage under separate ratios---so critic architecture and credit assignment move together and the gap cannot be assigned to the critic alone. They further differ in curriculum, action dimensionality, finger control and reward weights, the unified run ended on a task that commands standing rather than walking, and it is evaluated on a locomotion branch it did not train with.

\textbf{(2) Anti-gaming reward mechanisms did not raise validated reach rate.} In validated reach rate the dual-critic policy without anti-gaming mechanisms slightly outperforms the variant with five of them: 65.2\% vs.\ 60.9\% standing (S6s vs.\ S7) and 63.1\% vs.\ 56.2\% walking. S7 completed its own 8-level curriculum, and although it ran fewer iterations (15{,}000 vs.\ 20{,}000) its larger environment count means it collected roughly 50\% more environment steps, so a smaller sample budget is not an available explanation for the gap, although the two runs also differ in curriculum, environment count and arm initialization, and we did not measure which of these matters. S7 is in fact \emph{faster} to target than S6s in both modes (5.8 vs.\ 6.5 steps standing, 5.9 vs.\ 7.0 walking); it is validated reach rate and throughput on which it does not improve. We therefore report the absence of a gain on those two metrics, not an absence of effect.

\textbf{(3) The training curves cannot compare the two runs; only the standardized evaluation can.} The unified-critic run ended with training reward 36.2 against 37.1 for the dual-critic run, and accumulated 3.3M training reaches. Neither figure can be set against its counterpart: the reward tables differ, and the two runs count reaches under different definitions (raw position reaches for the unified run, validated reaches for the dual one), so we report no dual-critic reach total. The efficiency difference---3.5$\times$ in speed and 2$\times$ in throughput in the standing mode, 1.3$\times$ and 1.1$\times$ with the unified run driving its own fingers---is therefore measurable only through standardized evaluation with time-to-reach and throughput measurements. We likewise exclude curriculum level: once the two schedules are placed on their own scales (10 of 40 vs.\ 12 of 13), curriculum progression is \textit{not} comparable between the runs, and treating it as such would obscure a genuine difference rather than reveal one.

\textbf{Open question: RL fine-tuning of imitation-learned policies.} We did not train or evaluate an IL+RL pipeline, so what follows is a motivation for future work rather than a result. Consider a manipulation policy first learned from human demonstrations (e.g., via teleoperation) and then fine-tuned with RL to improve precision or adapt to new objects. With a shared critic, a locomotion reward that was never part of the imitation objective would enter the arm update; with separate critics, each reward would reach only its own branch's update. Whether that separation preserves imitation-learned behavior during RL refinement, a concern raised in recent IL+RL work~\cite{haldar2023teach}, is untested here.

\subsection{Confounding Factors}
\label{sec:confounds}

The comparison rests on a single pair of training runs, and those runs differ in more than the critic. We state what was and was not held constant.

\textbf{Held constant.} Robot and simulator; the same 17 policy-controlled body joints (the unified run additionally commanded seven finger joints, which are listed among the confounds below rather than here); observation spaces (57-dim locomotion, 52-dim arm); network widths; PPO settings (lr $3\times10^{-4}$ cosine, $\gamma=0.99$, $\lambda=0.95$, clip 0.2, 24-step rollouts); 2048 parallel environments; 20{,}000 training iterations; and the evaluation harness---one benchmark script, one locomotion branch given to all three policies (the dual run's, which S6u did not train with; see (vi) below), seed 42, and, in the original, pinned-finger evaluation, fingers pinned open for every policy. The arm reward is \emph{not} held constant: of its terms only the distance weight (4.0) is common to the two configurations, and the unified one carries a gripper term (weight 5.0) that was active for 18{,}837 of its 20{,}000 iterations, one fewer than the 18{,}838 iterations spent at Level 10 or above, because the reward term is first logged on the update following the level change.

\textbf{Not held constant.} (i)~\textit{Curriculum.} As described in Sec.~III-B the runs follow different schedules, and ended on tasks of very different difficulty---standing still versus walking at 0.6\,m/s with arbitrary commanded end-effector orientation. This is the largest of the confounds. (ii)~\textit{Arm action dimensionality.} S6u's arm actor emits 12 values (5 arm $+$ 7 finger) against S6s's 5. Finger control activates at Level 10 of that curriculum, so the seven finger outputs drove the robot for 18{,}838 of the 20{,}000 iterations, and they are discarded in the original, pinned-finger evaluation, where the fingers are pinned open for every policy. The unified policy is therefore evaluated there in a configuration it did not train in, and the confound lies in the task performed and may also lie in exploration and policy entropy, which were not measured here. (iii)~\textit{Reward weights.} The two configurations do not share a reward table. Forward velocity tracking is weighted 2.5 in the unified run against 5.0 in the dual one, active for the entire length of both runs, and the term lists themselves differ: the dual configuration adds leg-posture, standing-still and foot-stability terms, the unified one gripper, height-tracking and payload terms. (iv)~\textit{Seeds.} Neither training script set a random seed, so each arm is a single uncontrolled run with no variance estimate; the evaluation harness does seed itself, so the policies are compared on matched target sequences. (v)~\textit{Update rule.} The unified configuration normalizes one summed advantage, applies a single likelihood ratio to both actors, and clips gradients over the whole network with one optimizer. The dual configuration computes a separate advantage per reward stream, normalizes each, and updates each actor with its own ratio, its own gradient clipping and its own optimizer. Critic architecture and credit assignment therefore change together between the two runs. (vi)~\textit{Locomotion branch at evaluation.} The harness gives all three policies the locomotion branch the dual-critic runs share (Sec.~\ref{sec:setup}); S6u's own locomotion weights differ from it, so its arm is evaluated with a locomotion branch it did not train with. We have not measured what this costs, and the finger re-evaluation leaves it in place.

\textbf{Evidence against the training-progress explanation.} Honesty cuts both ways, and one observation resists the ``S6u simply progressed less far'' reading. The standing-mode benchmark uses a 0.06\,m position threshold, more lenient than the 0.05\,m S6u faced at its final level, and samples targets 0.18--0.40\,m from the shoulder, against 0.18--0.35\,m at that level and 0.18--0.40\,m at S6u's Levels~6--9. S6u is nonetheless slower to target in that mode. The size of that deficit depends on the harness: 3.5$\times$ with the fingers pinned; with S6u's own finger outputs restored, the standing-mode gap is 1.3$\times$ in speed and 1.1$\times$ in throughput. The effect of the locomotion-branch mismatch (vi) has not been measured. The walking mode inverts the reading one might expect: it is where S6u had trained least, and it is also where the corrected gap almost closes (62.5\% vs.\ 63.1\%, 13.3 vs.\ 13.7 reaches per 1{,}000 steps). We therefore state only this: the position threshold was looser than at S6u's final level, and the target range had been part of its curriculum, although the band beyond 0.35\,m was sampled only at Levels~5--9 (up to 0.38\,m at Level~5), in iterations 327--1{,}161 of 20{,}000; this weakens but does not eliminate the training-progress reading.

\textbf{The S6s--S7 comparison carries the same caveat.} It too is a single pair of runs, differing in iterations (15{,}000 vs.\ 20{,}000), environment count (4096 vs.\ 2048), curriculum and arm initialization; we read its null result with the same caution we apply to the main comparison.

\textbf{What this permits and forbids.} The evaluation is internally consistent---one harness, one shared locomotion branch, matched target sequences---so the measured gap is a real property of these two checkpoints \emph{under this harness}. It depends on the harness: with S6u's finger outputs restored and nothing else changed, the standing-mode gap falls from 3.5$\times$ to 1.3$\times$ in speed and from 2$\times$ to 1.1$\times$ in throughput. For the difference that remains, training progress is at least as parsimonious an explanation as critic architecture, and the unmeasured locomotion-branch mismatch (vi) is a third candidate; this design cannot isolate the critic's contribution from the other differences listed above. Establishing the causal claim requires a single-variable ablation---one curriculum, one action space, one reward set, one update rule, only the critic swapped, across several seeds---which we have instrumented but not yet run. The present result should be read as its starting point.

\section{Conclusion and Future Work}

We compared unified and dual critic architectures for humanoid loco-manipulation on the Unitree G1. Both runs learn functional reaching, but in the standing mode of our evaluation the dual-critic run reaches 3.5$\times$ faster with 2$\times$ higher throughput (1.3$\times$ and 1.1$\times$ with the unified run driving its own fingers), and five additional anti-gaming reward mechanisms did not raise validated reach rate. We also report a methodological observation: the two runs' training curves cannot be compared, since their reward tables and reach definitions differ, and only the standardized evaluation measures the difference, whose size depends on the evaluation itself (Sec.~\ref{sec:results}).

Our central claim is deliberately modest. Critic architecture is a design variable worth measuring rather than adopting by default, and we report a concrete efficiency gap between two runs while leaving its mechanism open. We do not claim to have isolated it. Limitations include simulation-only evaluation, single-run training with no seed set in either run, 5-DoF arm control, an evaluation protocol that pinned the unified run's fingers although it had trained driving them (restoring them shrinks the standing-mode gap from 3.5$\times$ to 1.3$\times$) and that ran it on a locomotion branch it did not train with, and---most importantly---the confounding of the update rule, curriculum schedule, arm action dimensionality, finger control and reward weights with the critic architecture (Sec.~\ref{sec:confounds}).

The immediate next step is therefore the single-variable ablation specified above: identical curriculum, action space, reward set and update rule, only the critic swapped, across at least three seeds. To make such comparisons auditable, every training run in our codebase now emits a machine-readable configuration record---all arguments, the fully resolved curriculum, every reward weight, observation and action dimensions, and the git commit---so that any two runs can be checked field by field before being called a comparison. Beyond that we plan to extend to 29-DoF dual-arm control with a Triple Actor-Critic, to test whether the per-branch credit assignment of the dual-critic design transfers to hybrid IL$+$RL pipelines as a mechanism for preserving imitation-learned skills, to integrate vision-language models for hierarchical task specification, and to pursue sim-to-real transfer.

\section*{Acknowledgment}

AI assistants (Claude, Anthropic) were used for code auditing, running evaluation scripts and drafting parts of the text, including the manuscript outline and, in this revision, the corrections in the abstract, Sections~I--V and Table~\ref{tab:results} and the added bibliography entries; all decisions and the final text are the author's responsibility.

\end{document}